# Ensuring Reliability of Curated EHR-Derived Data: The Validation of Accuracy for LLM/ML-Extracted Information and Data (VALID) Framework

[1]Melissa Estevez, MS; [2]Nisha Singh, MS; [2]Lauren Dyson, MS; [2]Blythe Adamson, PhD, MPH; [2]Qianyu Yuan, PhD; [2]Megan W. Hildner; [2]Erin Fidyk, MSN, MBA; [2]Olive Mbah, PhD, MHS; [2]Farhad Khan, MSPH; [3]Kathi Seidl-Rathkopf, PhD; [2]Aaron B. Cohen, MD MSCE

[1]Flatiron Health, Durham, NC, USA
[2]Flatiron Health, New York, NY, USA
[3]Flatiron Health Germany, Berlin, Germany

Corresponding Author:
Aaron B. Cohen, MD, MSCE
Flatiron Health
New York, NY, USA

Key Words: large language models (LLM), electronic health records (EHR), real world data (RWD), oncology, data quality, reliability

## Abstract

Large language models (LLMs) are increasingly used to extract clinical data from electronic health records (EHRs), offering significant improvements in scalability and efficiency for real-world data (RWD) curation in oncology. However, the adoption of LLMs introduces new challenges in ensuring the reliability, accuracy, and fairness of extracted data, which are essential for research, regulatory, and clinical applications. Existing quality assurance frameworks for RWD and artificial intelligence do not fully address the unique error modes and complexities associated with LLM-extracted data. In this paper, we propose a comprehensive framework for evaluating the quality of clinical data extracted by LLMs. The framework integrates variable-level performance benchmarking against expert human abstraction, automated verification checks for internal consistency and plausibility, and replication analyses comparing LLM-extracted data to human-abstracted datasets or external standards. This multidimensional approach enables the identification of variables most in need of improvement, systematic detection of latent errors, and confirmation of dataset fitness-for-purpose in real-world research. Additionally, the framework supports bias assessment by stratifying metrics across demographic subgroups. By providing a rigorous and transparent method for assessing LLM-extracted RWD, this framework advances industry standards and supports the trustworthy use of AI-powered evidence generation in oncology research and practice.

## Introduction

As stakeholders across life sciences, academia, and regulatory agencies increasingly turn to electronic health record (EHR)-derived data to support oncology research, regulatory submissions, and commercial strategy, the need for scalable, reliable extraction methods that produce high quality data has never been greater. The rapid evolution of artificial intelligence (AI)/machine learning (ML) and large language models (LLMs) in particular is transforming the landscape of real-world data (RWD) curation in oncology and making it possible to generate real world evidence (RWE) at greater scale. Yet, the very features that make LLMs so powerful, including their ability to process vast volumes of unstructured clinical text and generate structured variables at unprecedented speed, also introduce new complexities and risks. For researchers and institutions, the central question is no longer whether to use LLM-extracted data, but how to rigorously evaluate its reliability, fairness, and fitness for purpose in high-stakes applications.

One of the primary challenges is ensuring the reliability and accuracy of LLM-generated data, particularly in healthcare settings where data quality can profoundly impact clinical decisions and patient outcomes. LLMs can exhibit instability in generated datapoints, even when there are no changes to the prompt or inputs, leading to inconsistencies [1]. Additionally, these models are prone to hallucinations, producing information that appears plausible but is incorrect or fabricated [1,2]. Finally, LLMs may struggle to handle the clinical nuances and subjectivity present in clinical documentation as effectively as a human expert [3]. While open source LLMs trained on EHR data are emerging, most commonly used open-source LLMs are not pre-trained on EHR data and may lack the clinical domain knowledge needed to perform well on the tasks where EHR documentation is inconsistent, missing, or subject to interpretation [4]. The complexity of medical language and the variability in clinical documentation further complicate the extraction process, necessitating a robust framework to assess the quality of the data produced by these models.

A growing body of literature and regulatory guidance has sought to address the challenges of quality assurance in RWD and AI applications. The U.S. Food and Drug Administration (FDA) recently issued guidance emphasizing the need for transparency, traceability, and rigorous validation of AI/ML-based tools used in clinical research and healthcare decision-making [5]. The FDA's 2025 framework highlights the importance of data provenance, model versioning, and continuous monitoring, but stops short of providing methodologies for evaluating the unique error modes and performance characteristics of LLMs applied to unstructured EHR data. Similarly, the International Society for Pharmacoeconomics and Outcomes Research (ISPOR) has published good practices reports and checklists, such as the PALISADE Checklist for machine learning in health economics and outcomes research, which offer valuable recommendations for transparency, reproducibility, and bias assessment in traditional machine learning models and the SUITABILITY Checklist for assessing the quality of real-world data from EHRs [6,7].

However, these frameworks were developed primarily for structured data and conventional ML algorithms, and do not provide detailed recommendations around implementing accuracy assessments. Prior studies have demonstrated LLM/ML-based methods can be used to reliably extract variables from the EHR and propose approaches to accuracy assessment [8,9]. However, current frameworks and studies do not fully address the challenge of benchmarking against human abstraction at scale or the complexities introduced when LLMs are used across multiple variables to generate entire datasets or to answer a research question. Other published frameworks, including those from the Coalition for Health AI and the Holistic Evaluation of Language Models (HELM) project, provide important dimensions for model evaluation, such as fairness, robustness, and calibration, but lack practical guidance for operationalizing these metrics in the context of clinical variable extraction from EHRs [10]. As a result, a critical gap remains in the literature: no existing framework offers a comprehensive approach for assessing the accuracy and reliability of LLM-extracted clinical data from the EHR, leaving users of the data (eg, life science companies, academic researchers, governing bodies) without clear standards for dataset evaluation or internal quality assurance.

This paper aims to move beyond existing limited resources and provide a practical, transparent framework for evaluating the quality of model-extracted oncology data, including data extracted by LLMs and other ML models. Drawing on the experience of Flatiron Health in developing and extensively validating ML- and LLM-based curation pipelines [11–24], and building on past approaches for evaluation of LLM/ML-extracted data [25], we outline a holistic approach to assess accuracy and completeness of data extraction approaches that includes evaluation across multiple LLM/ML-extracted variables in a dataset. While other components of data quality, such as relevance (including availability, sufficiency, and representativeness) and other aspects of reliability (provenance, timeliness), are important, they are out of scope for this paper and have been previously described [26]. In addition, the framework focuses on the accuracy and completeness of data extracted by ML/LLMs relative to what is documented in the source EHR; however, evaluating the accuracy and completeness of the source data itself for the research question is still needed. By providing a framework to interrogate data quality at every stage, from model development to dataset delivery, we seek to raise the industry standard and foster greater confidence in the use of AI-powered RWD for oncology research and decision-making.

## Validation of Accuracy for LLM/ML-Extracted Information and Data (VALID) Framework

The accuracy assessment framework presented here is designed to provide a comprehensive, transparent, and reproducible approach for evaluating the quality and fitness-for-purpose of clinical data extracted from EHRs using LLMs or other ML models. Recognizing the unique challenges and opportunities posed by LLMs, including their ability to process unstructured text at scale and their potential for novel error modes, this framework moves beyond traditional accuracy metrics to offer a holistic evaluation strategy. The framework is built on three foundational pillars: variable-level performance metrics, automated verification checks, and replication and benchmarking analyses (Figure 1).

By integrating these three components, the VALID framework provides a multidimensional view of LLM-extracted data quality. This approach not only quantifies accuracy but also surfaces latent errors and bias, supports continuous model improvement, and builds confidence in the use of LLM-extracted RWD for oncology research and decision-making. Below we go into greater detail about each section of the accuracy assessment framework.

### 1 Variable-Level Performance Metrics

Variable-level performance metrics measure the accuracy and completeness of a data curation approach by comparing it against a source of truth label. These metrics include standard accuracy measures such as recall, precision, F1, as well as completeness rates, which assess the percentage of patients with a known value. EHR data is inherently limited to the data captured within a specific health practice or network, which may not represent the full scope of a patient's care. As a result, these metrics evaluate the accuracy and completeness of LLM-extracted data relative to what is available in the source EHR documentation. Performance metrics are generated using a held-out testing data set that is not used for model training, is representative of the target population, and is sufficiently large to ensure statistical reliability. During the development of a new LLM approach, model performance can also be assessed using a prototyping test set (i.e. a dataset that is used to iterate on the model-extraction approach, including pre-processing and LLM prompts).

This framework assesses LLM performance by comparing both LLM-extracted and expert human-abstracted data against a common reference standard (i.e., the ground truth that the LLM and single human abstractor is compared to) (Table 1). Even when there are rigorous quality control measures for expert human abstraction (including standardized procedures and inter-rater agreement monitoring), some disagreement among abstractors is inevitable due to the inherent complexity and ambiguity of documentation in EHRs (e.g., conflicting, incomplete or unclear documentation that could reasonably be interpreted two different ways for a nuanced concept such as date of cancer progression). The methodology, demonstrated with illustrative examples in Table 2, calculates both absolute performance metrics and the relative performance differences between LLMs and human abstractors. This relative performance measure (the difference between LLM and human accuracy metrics) provides more consistent comparisons across different variables and datasets by accounting for the natural variation in human performance. This relative measure also offers a more practical and understandable benchmark to determine when a model's quality is sufficient for the research purpose or if further iteration is needed, relative to setting thresholds for absolute metrics.

The approach of comparing LLM performance relative to expert human abstraction enhances accuracy metric interpretation through human-model comparison. By measuring abstractor disagreement rates, we can identify when LLM performance appears low due to human labeling challenges rather than model limitations. Abstractor performance also reveals the inherent complexity or subjectivity of clinical concepts being extracted. For example, Table 2 provides a comparison of two variables, surgery and locoregional recurrence, with similar model performance against single-abstracted labels. However, when examining expert human abstraction metrics, we discover that surgery data has higher human consensus than locoregional diagnosis data. This indicates locoregional diagnosis may be inherently more complex or subjective in a patient's record, which introduces ambiguity that increases the difficulty of the task and results in varying levels of performance for both human abstractors and models. Evaluating the relative performance difference between humans and LLMs (-10% pt for surgery vs. -5% pt for locoregional recurrence) helps contextualize model performance relative to task difficulty, highlights where to focus model development resources for the greatest potential gains, and reveals opportunities where LLMs may even exceed human-level performance.

End-to-end evaluation metrics, as shown for triple negative breast cancer (TNBC) status at metastatic breast cancer (mBC) diagnosis in the last row of Table 2, are essential for complex derived variables such as line of therapy or biomarker status at key timepoints. Using TNBC status at mBC diagnosis as an example, accurate patient subgrouping requires the correct alignment of multiple extracted variables including: the test date and result for estrogen receptor (ER), progesterone receptor (PR), human epidermal growth factor receptor 2 (HER2) as well as the mBC diagnosis date. Even with good performance on individual variables, compounding errors can significantly reduce end-to-end accuracy. These comprehensive metrics provide more clinically and analytically relevant performance indicators. Creating abstraction performance benchmarks for these complex metrics requires adjudication or duplicate abstraction for all component variables across the same patient cohort (Table 1). However as with individual variables, abstractor benchmarks are critical for interpreting performance of these derived variables.

Finally, while human benchmarks are useful for interpreting and prioritizing quality improvement work, they may not be needed for every model. Some examples include a simple clinical concept where human abstractors are likely to have high agreement (e.g., history of smoking: yes/no), or situations where LLM/ML performance on single abstracted labels is very high (e.g., model F1: 99%). The intended use case may also influence the utility of human benchmarks (e.g., regulatory use case versus hypothesis generation).

## 2 Automated Verification Checks

Verification checks assess the prevalence of conflicting or erroneous data points at the patient level and ensure that cohort-level distributions align with clinical expectations. These checks serve as a proxy for accuracy and help evaluate the face validity and usability of the dataset. Even when LLM extraction performance is high, small amounts of misclassified data may result in usability issues as well as general distrust in the dataset. Verification checks help mitigate this issue by surfacing data inconsistencies that are likely the result of model errors; resolving them improves the overall quality and usability of the dataset.

Previously described verification checks provide a powerful tool to assess quality of LLM-extracted data [26]. These checks fall into three categories: conformance, plausibility and consistency. Table 3 provides examples of checks performed to assess a fully LLM-extracted breast cancer dataset.

Clinicians play a pivotal role in developing and interpreting a robust and comprehensive set of checks and identifying which inconsistencies require further investigation. This process allows for the targeted resolution of errors and results in a cleaner and more accurate dataset. Some conflicts, such as a metastatic diagnosis date prior to initial diagnosis, are clinically illogical and likely reflect a model error. It is clear that these discrepancies should be investigated and addressed. However, other conflicts may reflect real-world scenarios. For example, a patient receiving targeted therapy for a mutation they appear to lack in the EHR may be a result of incomplete biomarker data. It also could be off-label use and a true medical finding. A clinical team partnering with engineers to review the source documentation (if possible) can help tease this out. Not only does this ensure accurate answers, but it also gives insight into why the errors are occurring in the first place to inform any changes in our models that may be needed.

## 3 Replication and Benchmarking Analyses

Replication and benchmarking analyses assess whether an analysis performed using LLM-extracted data produces similar results and conclusions when compared to a reference dataset. The reference dataset can be an internal benchmark, such as a dataset built using expert human abstraction, or an external benchmark, such as a comparative treatment effect or prognostic marker that has been well-established in literature or an external dataset such as the Surveillance, Epidemiology, and End Results (SEER) database. While variable metrics and verification checks focus on one, or at most a few, variables at a time, replication and benchmarking analyses utilize the full dataset to answer a research question. This gives replication and benchmarking analyses the ability to describe how model errors across multiple LLM-extracted variables may interact together and potentially introduce bias when selecting cohorts and assessing outcomes. As a result, replication analyses are critical for assessing the fitness-for-purpose of RWD in research and regulatory contexts [13].

Like with verification checks, selection and design of replication and benchmarking analyses should be clinically informed and consider a multitude of RWD applications. Based on common use cases for disease-specific RWD cohorts, we propose conducting analyses in both: (1) broad cohorts, and (2) specific subcohorts of interest (e.g., biomarker or treatment based).

### 3.1 Broad Cohort Characterization

Broad cohort characterization validates outcomes and clinical distributions relative to an internal reference standard in a broad cohort. The definition of broad cohorts will vary by disease, but will most likely be based on disease setting (e.g., early vs. advanced/metastatic) alone or in combination with disease subtype (e.g., hormone sensitive vs. castrate resistant in prostate cancer). Since cancer care and outcomes vary significantly by disease setting and subtype, assessing outcomes and cohort distributions by disease setting is more interpretable than across a larger population. Because these analyses are assessing broad populations, rather than niche biomarker or treatment defined subgroups, obtaining an internal reference standard may be feasible, but cohort-level distributions and outcomes can also be assessed relative to clinical expectations, treatment guidelines, and the literature.

These analyses consist of several types of outputs: cohort distributions, trends in clinical care over time, and patient outcomes. Cohort distributions assess the distribution of demographic and clinical characteristics (e.g., age at diagnosis, gender, distribution of stage, histology, menopausal status), biomarker testing information (e.g., testing and result rates, time from diagnosis to first test, biomarker status at key index dates) and treatment information (e.g., surgery/radiation rates stratified by stage, proportion of patients receiving neoadjuvant therapy, most prevalent 1L treatment regimens stratified by breast cancer subtype). Trends in clinical care over time can validate whether the data exhibits expected patterns, such as an increase in first-line treatment starts for a new drug that was recently approved. Finally, these reports validate meaningful patient outcomes by disease setting, such as recurrence-free survival (RFS) and overall survival (OS).

### 3.2 Subcohorts of Interest

Replication and benchmarking at the subcohort level is used to understand the reproducibility of analytic results and scientific conclusions for subpopulations of interest. While broad cohort characterization is important for validating the dataset as a whole, it has a similar limitation as test set metrics in that it may not detect where performance may differ for a specific subcohort. For example, consider an LLM that extracts oral treatment from the EHR. The model may have high performance on the broad cohort, but model errors may also be disproportionately high for patients who discontinue and then restart the oral treatment, resulting in multiple treatment start and stop dates for the same oral drug. If multiple treatment spans happen more commonly in certain subpopulations (e.g., older patients) this can result in differential errors in this patient cohort.

To ensure replication and benchmarking analyses are clinically meaningful and aligned with real-world practice, clinician input is essential in defining and prioritizing subgroups of interest. Disease experts play a central role in identifying these cohorts based on clinical guidelines (e.g., National Comprehensive Cancer Network [NCCN], American Society of Clinical Oncology [ASCO]), therapeutic relevance, and RWD study needs. Cohort definitions are anchored in key clinical components, including disease setting (e.g., early vs. metastatic), histology or subtype classifications (e.g., hormone receptor [HR]/HER2 status), biomarker profiles (e.g., Breast Cancer gene [BRCA] status, programmed death ligand 1 [PD-L1] expression), treatment setting (e.g., adjuvant, first-line metastatic), and other relevant inclusion/exclusion criteria. Clinicians also guide the selection of index dates and outcome measures to ensure consistency, comparability, and clinical relevance across studies.

Once subcohorts are selected, the next consideration is what reference standard is available. If the subcohort is a highly prevalent population, such as HR+/HER2- breast cancer, it may be possible to assess outcomes like treatment patterns and OS relative to an internal reference standard (e.g., a human abstracted dataset). However, if replicating outcomes and treatment patterns for a smaller subpopulation, (e.g., patients with BRCA mutations or patients with HR+/HER2- taking a newly approved first-line regimen) obtaining reference data may not be feasible and external benchmarking may be required. When benchmarking to literature, it may be the case that the specific point estimates for the outcome (e.g., median OS) do not replicate due to population differences (i.e. real-world vs. clinical study cohorts). However, we may still expect that similar conclusions can be drawn (e.g., an improved median OS of one treatment over another is still preserved despite differences in the actual OS point estimates).

## Applying VALID Framework for Bias Assessment

Each of the accuracy components described in the VALID framework can be leveraged to provide a comprehensive assessment of model bias. Even when models are high performing overall, there is a potential risk that performance could vary across clinical and demographic subgroups, leading to biased RWD. If bias is not assessed and addressed, this could lead to misleading or inaccurate conclusions when RWD is used to generate RWE and can perpetuate existing inequities [27,28].

First, the approach of evaluating variable level performance metrics and evaluating the relative difference between humans and LLMs can be extended to assess model bias by stratifying metrics by demographic subgroups (e.g., race/ethnicity, age, gender) or clinical subgroups (e.g., by treatment, biomarker status, stage). Again, obtaining metrics for both the model and abstraction for these subgroups offers a unique advantage. If differences in performance are observed in only the model metrics, but not the abstractor metrics, it indicates that model errors may be differential (i.e., occur more frequently in certain subgroups). When differences in performance are observed in both model and abstractor metrics, it may signal that more challenging and complex cases are not evenly distributed within the dataset. In this situation, reviewing cases and identifying error modes may lead to insights on how to improve quality.

Model bias can also be assessed by stratifying verification checks by demographic subgroups or key subpopulations of interest. Since data conflicts can serve as a proxy for quality, stratified verification checks can identify model bias and ensure that certain subgroups are not at risk for being disproportionately removed from analyses due to non-sensical data. For variables with rare classes and those with less prevalent demographic subgroups, testing data sets may be too small to generate stratified variable-level metrics. Verification checks offer an alternative since they do not require a reference standard and are applied to the full dataset.

Finally, replication and benchmarking can be used to assess model bias by replicating a health equity research question. For example, can LLM-generated data replicate findings that are well established in the literature, such as lower survival among Black patients with TNBC compared to their White counterparts? Simple replication analyses using an internal reference standard (e.g., OS patterns of patients with mBC) can be stratified by demographic subgroups or key subpopulations of interest when internal reference standards are sufficiently large.

## Discussion

We present a comprehensive accuracy assessment framework for evaluating LLM-extracted data from the EHR. This approach, which is model agnostic and also applies to traditional ML extraction approaches, extends beyond conventional performance metrics to deliver a transparent and reproducible evaluation of oncology EHR LLM-extracted data quality. Central to this framework is the rigorous benchmarking of LLM accuracy metrics such as recall, precision, and completeness, to human-level performance. By calculating the relative difference between LLM and human performance rather than focusing solely on absolute metrics, especially for more complex clinical concepts or variables that may be used for regulatory or other high stakes applications, RWD developers can better contextualize model performance relative to humans, prioritize resource allocation for model improvements, and understand when LLM performance may exceed that of a human abstractor. Automated verification checks further enhance quality assurance by systematically flagging internal inconsistencies, implausible values, and temporal conflicts that may elude variable-level metrics, thereby enabling early detection and remediation of latent errors. The third pillar, replication and benchmarking analyses, evaluates the ability of LLM-extracted data to reproduce established clinical distributions and outcomes, either relative to internal human-abstracted datasets or external epidemiological benchmarks, thus providing critical evidence of fitness-for-purpose in research and regulatory settings. Replication helps build trust in LLM-extracted data when results are consistent with expectations and also identifies use cases or subpopulations where analyses are not concordant and can be used to prioritize further quality improvement efforts and analytical guidance. Finally, by integrating these multidimensional assessments, the framework not only quantifies accuracy, but can also be applied to assess model bias. This holistic approach is essential for ensuring that the promise of LLM-driven data curation translates into robust, equitable, and actionable real-world evidence in clinical practice and research.

Local context must be considered across all three pillars of the framework when evaluating LLM-based extraction in different countries or treatment settings. Documentation patterns or EHR systems, for instance, may vary in a way that makes extracting certain information systematically more difficult in some countries or languages than others. Evaluating LLM-based extraction against local abstraction metrics is essential for accurately contextualizing differences in model performance across different settings. Similarly, patterns of care, including testing frequency or treatment guidelines, can vary across countries. These differences need to be taken into account when constructing automated verification checks such as plausibility checks and in choosing adequate reference standards for conducting replication analyses.

The effective application of this framework demands a multidisciplinary team with diverse expertise. Clinical teams are essential for understanding and identifying use cases and their feasibility with RWD, designing meaningful verification checks, choosing appropriate subcohorts for replication assessment, interpreting the clinical significance of analysis results, and investigating chart-level errors for insights that can inform prompt refinement. Research scientists and health equity experts provide crucial context regarding the impact of model errors on downstream research applications, ensuring that quality assessments and quality improvement efforts remain focused on errors with meaningful consequences for evidence generation. Engineers develop and help maintain the technical infrastructure needed to implement metrics and verification checks in an efficient and scalable manner, and iterate on models when needed to optimize performance (e.g., prompt engineering, fine-tuning). Without this collaborative approach, the framework may focus improvement efforts on issues with limited research impact or may fail to detect clinically significant errors altogether, ultimately resulting in lower data quality and potentially incorrect conclusions that may impact research or patient care.

Despite its comprehensive approach, our framework faces several practical limitations that require careful consideration during implementation. First, effective evaluation of model performance against human abstraction is only meaningful when the reference data is of known and sufficient quality. Without this, poor-quality reference data can confound performance assessments, leading to misleading conclusions and false reassurance about both model accuracy and overall data quality. The higher the quality of the reference data, the more interpretable and trustworthy the resulting performance metrics will be. Second, the framework requires pragmatic balancing of scientific rigor with realistic constraints. It may not always be possible to generate duplicate abstracted or adjudicated data due to limitations in time or resources. Further, there are an endless number of verification checks that can be designed and subcohorts that can be investigated via replication, and eventually some of these assessments may become redundant or not lead to any meaningful change in the data. A risk-based approach is therefore necessary, where single abstracted data may be sufficient for variables with high performance metrics. These variables might also require fewer replication analyses and verification checks. Third, the dynamic nature of LLM capabilities introduces temporal challenges, as models continually evolve through version updates and fine-tuning. Each significant change to models or data delivery pipelines may necessitate re-running quality assessment analyses to ensure consistent performance, creating an ongoing maintenance burden that must be factored into operational planning. Finally, it is important to recognize that our framework primarily measures accuracy relative to what is documented in the EHR, leaving the fundamental issue of missingness and potential bias in source documentation. The most sophisticated extraction approach cannot recover information that was never recorded, highlighting the need for complementary assessment of other quality dimensions such as relevance, completeness, and representativeness of the underlying EHR data. These limitations underscore that while our framework represents a significant advancement in quality assessment for LLM-extracted data, it must be implemented with careful consideration of available resources, organizational capabilities, and the specific research contexts in which the extracted data will ultimately be applied.

While the proposed framework provides one approach to evaluating the accuracy and reliability of ML- and LLM-extracted EHR data, additional work is still required, including a need to build consensus across the field regarding what thresholds of LLM performance (as well as the reference data) are considered "good enough" for different research and regulatory applications. Comparative studies that benchmark LLM performance against expert human abstraction across diverse clinical variables and settings will be critical to inform these standards. In addition, as LLMs and EHR documentation practices continue to evolve, ongoing research is needed to ensure that quality assessment frameworks remain relevant and effective. This includes evaluating the impact of new model architectures, training data sources, and prompt engineering strategies on extraction quality. Finally, there is an opportunity to extend this framework beyond oncology to other diseases, therapeutic areas and healthcare systems.

## Conclusion

The integration of ML and LLMs into healthcare data extraction represents a transformative opportunity to accelerate and enhance RWD generation. Our accuracy assessment framework provides a structured approach to evaluate the reliability and quality of LLM-extracted data, addressing critical concerns about model performance across diverse patient populations and the unique error modes for LLMs such as hallucinations. By combining rigorous benchmarking against human abstraction, automated verification checks, and replication analyses, we have created a multidimensional assessment strategy that not only quantifies accuracy but also identifies biases, inconsistencies, and areas for model improvement. This framework supports a responsible path forward for the adoption of LLM technologies in oncology research, ensuring that efficiency gains do not come at the expense of data quality or patient representation. As healthcare systems increasingly adopt AI-driven approaches to data extraction, frameworks like this will be essential to maintain scientific integrity and foster trustworthy, consistent evidence generation. Establishing robust quality standards now lays the groundwork for responsible innovation that benefits researchers, clinicians, and patients.

**Acknowledgements**

Darren Johnson for support in publication planning and management. The authors used Dashworks to aid in the formatting of this manuscript for the first draft and take full responsibility for the content of this publication.

**Conflicts of Interest**

During the drafting of the work, ME, NS, LD, BA, QY, MWH, EF, OM, FK, KSR, and ABC reported employment with Flatiron Health, Inc. and stock ownership in Roche.

**Funding**

This study was sponsored by Flatiron Health, Inc.—an independent member of the Roche Group.

**Institutional Review Board Statement**

Not applicable

**Table 1.** Approaches for creating reference standards

| Approach | Description | Application and Interpretation |
|---|---|---|
| Duplicate Abstraction | For every patient in the test set, data elements are generated by the LLM as well as two expert human abstractors assigned to every task. The reference standard is the data elements generated by Abstractor 2. The performance of Abstractor 1 and the LLM are both measured against the reference standard (Abstractor 2). While we refer to 'Abstractor 1' and 'Abstractor 2', abstractors are randomly assigned to tasks and these do not refer to a single individual. | <ul><li>Useful during model iteration since the labels do not need to be updated as the model is changed</li><li>Absolute performance of both the LLM and abstractor will be underestimated since performance is being measured on single abstracted labels</li></ul> |
| Double Adjudication | For every patient in the test set, data elements are generated by the LLM as well as one abstractor assigned to every task. The reference standard is generated by having an expert human abstractor adjudicate every case of LLM <> abstractor disagreement. Single abstracted labels and the LLM are both evaluated against the reference standard. | <ul><li>Useful when model iteration is complete because changes in the model will surface new patients with LLM <> abstractor disagreement that require adjudication. Most resource efficient approach.</li><li>Absolute metrics will be more accurate than duplicate abstraction, but may be underestimated in the case where the model and abstractor agree on the wrong answer. However this approach can show where the LLM is outperforming expert human abstraction</li></ul> |
| Triple Adjudication | For every patient in the test set, data elements are generated by the LLM as well as two expert human abstractors assigned to every task. The reference standard is generated by having an expert human abstractor adjudicate every case of LLM <> abstractor disagreement and abstractor <> abstractor disagreement. Single abstracted labels and the LLM are both evaluated against the reference standard. | <ul><li>Useful when model iteration is complete because changes in the model will surface new patients with LLM <> abstractor disagreement that require adjudication.</li><li>This is the most rigorous approach but also the most resource intensive. Therefore it is useful only in cases where most accurate absolute metrics are needed</li></ul> |

Abbreviations: LLM, large language model.

**Table 2.** Examples with interpretation of LLM performance metrics to assess data quality

| Variable | LLM Performance | Abstraction Performance | Relative Performance Difference (LLM - Abstraction) | Interpretation |
|---|---|---|---|---|
| Surgery and date | 85% recall<br>85% precision<br>85% date accuracy | 95% recall<br>95% precision<br>95% date accuracy | -10% pt recall<br>-10% pt precision<br>-10% pt date accuracy | Model is further from human-level performance on what is likely to be a simpler clinical concept. Additional model development should be considered. |
| Locoregional recurrence and date | 85% recall<br>85% precision<br>85% date accuracy | 90% recall<br>90% precision<br>90% date accuracy | -5% pt recall<br>-5% pt precision<br>-5% pt date accuracy | Model is close to human-level performance on a more complex and ambiguous concept |
| Triple negative breast cancer (TNBC) status within +/- 60 days of metastatic diagnosis | 92% recall<br>90% precision | 90% recall<br>93% precision | +2% pt recall<br>-3% pt precision | Model is close to (and possibly exceeding) human performance on a clinically important cohort derived across multiple variables (ER/PR/HER2 test results and dates, and metastatic diagnosis and date) |

Abbreviations: ER, estrogen receptor; HER2, human estrogen receptor 2; LLM, large language model; % pt, percentage point; PR, progesterone receptor.

**Table 3.** Types of verification checks to assess the face validity of LLM-extracted RWD

| Category | Example Check | Interpretation |
|---|---|---|
| Patient level checks | Patients that have an initial breast cancer diagnosis that is equivalent to the metastatic diagnosis date should also show up as having stage IV disease | Conflicting data points likely reflect a model extraction error, or less likely, a documentation error |
| | Events occur in an expected order (e.g., surgery date is after initial diagnosis date, but before metastatic diagnosis date) | |
| | Patients should not have both a positive and negative gBRCA 1 result | |
| | HER2 result dates do not pre-date the development of HER2 tests | |
| | Events occur in an expected order (e.g., radiation does not occur prior to surgery for patients with breast cancer) | While these findings may be examples of real world practice or off label usage, patients flagged by these checks may also be enriched for model errors. |
| | Time between surgery and start of adjuvant therapy is within an expected range | |
| | Patients receiving endocrine therapy have positive HR status | |
| | The metastatic diagnosis date extracted by the LLM for a given patient doesn't change when the database is refreshed month over month in absence to a change in the underlying LLM | |
| Cohort level checks | Distribution of first-line treatment regimens for patients with TNBC reflect expected clinical practice as described by NCCN guidelines | Cohort distributions that yield unexpected results may identify models that are underperforming in specific subgroups |
| | Rates of surgery stratified by stage at initial diagnosis align with clinical practice as described by NCCN guidelines | |
| | The number of patients with a metastatic diagnosis date each month is consistent over time | |

Abbreviations: ER, estrogen receptor; gBRCA, germline BRCA; HER2, human estrogen receptor 2; HR, hormone receptor; LLM, large language model; NCCN, National Comprehensive Cancer Network; PR, progesterone receptor; RWD, real-world data; TNBC, triple-negative breast cancer.

**Figure 1.** Components of the EHR LLM-extracted RWD Quality Framework

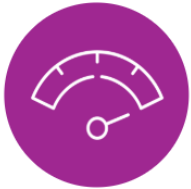
**Variable-Level Performance Metrics**

Performance *relative to expert human abstraction* for variable level and end-to-end metrics

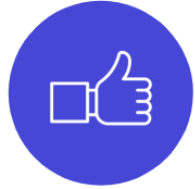
**Automated Verification Checks**

Patient- and cohort-level checks for face validity (e.g., obviously erroneous or unlikely data points)

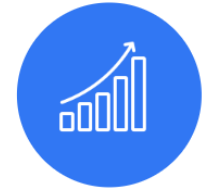
**Replication and Benchmarking Analyses**

Analytic results that align with internal (expert human abstraction) or external benchmarks